\documentclass[times,twocolumn,final]{elsarticle}

\usepackage{medima}
\usepackage{framed,multirow}
\usepackage{amsmath}
\usepackage{multirow}

\usepackage{amssymb}
\usepackage{latexsym}

\usepackage{url}
\usepackage{xcolor}

\usepackage{hyperref}

\definecolor{newcolor}{rgb}{.8,.349,.1}

\journal{Medical Image Analysis}

\begin{document}

\verso{Gen Shi \textit{et~al.}}

\begin{frontmatter}

\title{Benefit from public unlabeled data: A Frangi filtering-based pretraining network for 3D cerebrovascular segmentation}%

\author[1]{Gen \snm{Shi}}
\author[2]{Hao \snm{Lu}}
\author[3]{Hui \snm{Hui}\corref{cor1}}
\ead{hui.hui@ia.ac.cn}
\author[1]{Jie \snm{Tian}\corref{cor1}}
\ead{tian@ieee.org}
\cortext[cor1]{Corresponding author}

\address[1]{School of Engineering Medicine and School of Biological Science and Medical Engineering, Beihang University, Beijing, 100191, China, and also with the Key Laboratory of Big DataBased Precision Medicine (Beihang University), Ministry of Industry and Information Technology of China, Beijing, 100191, China}
\address[2]{State Key Laboratory for Management and Control of Complex Systems, Institute of Automation, Chinese Academic of Science, Beijing 10086, China}
\address[3]{CAS Key Laboratory of Molecular Imaging, Institute of Automation, Chinese Academy of Sciences, Beijing, 100190, China}

\received{1 May 2013}
\finalform{10 May 2013}
\accepted{13 May 2013}
\availableonline{15 May 2013}
\communicated{S. Sarkar}

\begin{abstract}
The precise cerebrovascular segmentation in time-of-flight magnetic resonance angiography (TOF-MRA) data is crucial for clinically computer-aided diagnosis. However, the sparse distribution of cerebrovascular structures in TOF-MRA results in an exceedingly high cost for manual data labeling. The use of unlabeled TOF-MRA data holds the potential to enhance model performance significantly. In this study, we construct the largest preprocessed unlabeled TOF-MRA datasets (1510 subjects) to date. We also provide three additional labeled datasets totaling 113 subjects. Furthermore, we propose a simple yet effective pertraining strategy based on Frangi filtering, known for enhancing vessel-like structures, to fully leverage the unlabeled data for 3D cerebrovascular segmentation. Specifically, we develop a Frangi filtering-based preprocessing workflow to handle the large-scale unlabeled dataset, and a multi-task pretraining strategy is proposed to effectively utilize the preprocessed data. By employing this approach, we maximize the knowledge gained from the unlabeled data. The pretrained model is evaluated on four cerebrovascular segmentation datasets. The results have demonstrated the superior performance of our model, with an improvement of approximately 3\% compared to state-of-the-art semi- and self-supervised methods. Furthermore, the ablation studies also demonstrate the generalizability and effectiveness of the pretraining method regarding the backbone structures. The code and data have been open source at: \url{https://github.com/shigen-StoneRoot/FFPN}.
\end{abstract}

\begin{keyword}
\MSC 41A05\sep 41A10\sep 65D05\sep 65D17
\KWD Keyword1\sep Keyword2\sep Keyword3
\end{keyword}

\end{frontmatter}


\begin{figure*}
\centering
\includegraphics[width=\linewidth]{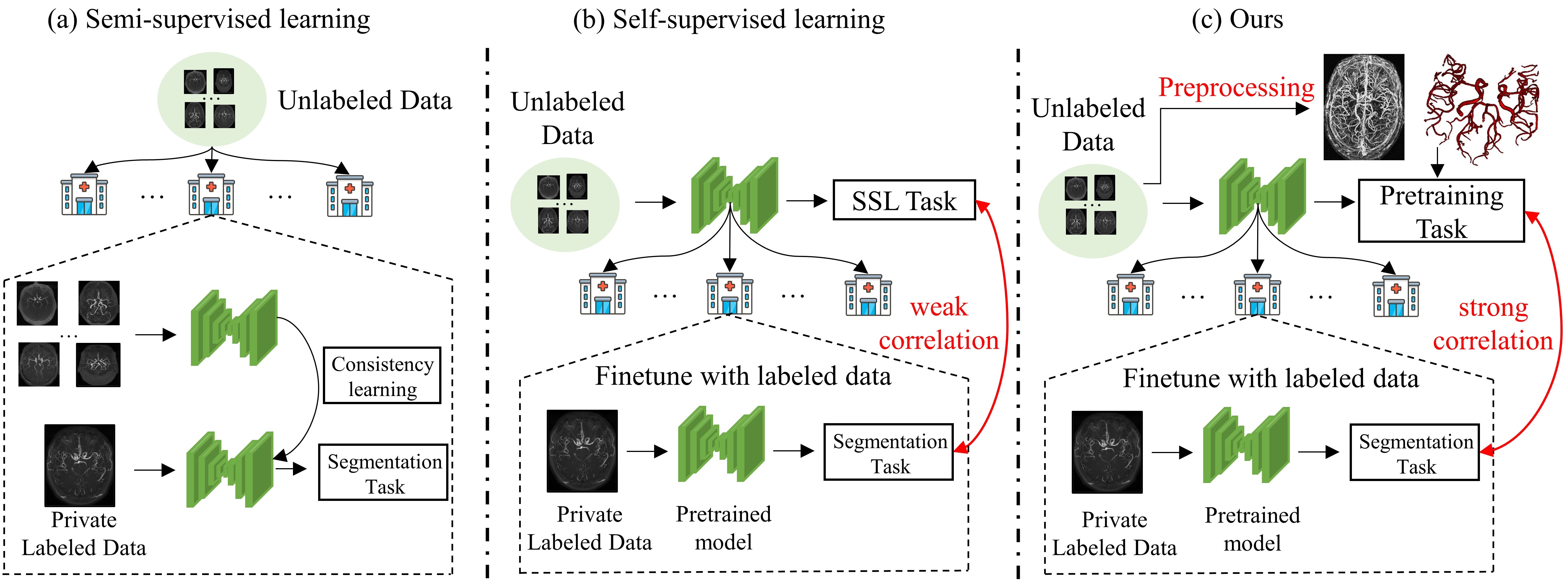}
\caption{The illustration of using unlabeled MRA-TOF data through (a) semi-supervised learning, (b) self-supervised learning and (c) our pretraining method.}
\label{fig:SemiSL_SSL}
\end{figure*}

\section{Introduction}
TOF-MRA is one non-invasive medical imaging technique to visualize cerebral blood vessels. It provides detailed images of the blood vessels in the brain, enabling early detection and treatment of potentially life-threatening conditions \cite{ozsarlak2004mr, hassouna2006cerebrovascular}. The accurate vessel segmentation is a crucial preprocessing step in TOF-MRA image analysis, which provides surgeons with essential information about the location, size, and connectivity of blood vessels and facilitates surgical planning and intervention \cite{ni2020global}.

Compared to the segmentation of other biological tissues, brain vessel segmentation based on TOF-MRA presents more challenges. Most organs typically have a spherical shape and are distributed in a concentrated manner, while blood vessels have a tubular structure and their distribution in TOF-MRA is relatively sparse \cite{chen2022generative, xia20223d}, which increases the difficulty and cost in obtaining manually labeled MRA-TOF data. On the contrary, the unlabeled TOF-MRA data are easily accessible from the public resource (i.e., IXI and OASIS3). Therefore, how to fully use the public unlabeled data is essential for cerebrovascular segmentation with limited manually labeled data.

There are two fundamental deep learning-based methods that are capable of using unlabeled data----semi-supervised learning (SemiSL) \cite{qi2020small, yang2022survey} and self-supervised learning (SSL) \cite{9770283, huang2023self} methods (see Fig.\ref{fig:SemiSL_SSL}~a and b). SemiSL-based methods typically utilize unlabeled data in conjunction with labeled data during the training process, learning the consistency between them \cite{cheplygina2019not}. This process requires repeated access to the unlabeled data and increases the demand for computational resources, when different research institutes want to use these data. Besides, the public unlabeled data are collected from multiple resources and may be highly heterogeneous from the single-site dataset. The issue of distribution shift, arising from this, has the potential to adversely affect the performance of SemiSL methods \cite{chen2019distributionally}. Regarding the SSL methods, they usually pretrain the model through the pretext task \cite{zhang2023dive}, and then the pretrained model can be reused by different institutes without access to the unlabeled data. This process decreases the computational cost when the unlabeled data scale is large and protects patient privacy since different institutes do not directly access the unlabeled data \cite{asadian2022self}.

However, the performance of SSL methods highly depends on the choice of pretext task. The pretext tasks, such as contrastive learning between augmented data \cite{chen2020simple, liao2022muscle} and masked imaging modelling \cite{he2022masked}, are significantly different from the downstream task (e.g., cerebrovascular segmentation in this study). This setting proves advantageous when dealing with multiple downstream tasks, while a specific single downstream task might not gain much benefits from SSL methods. Therefore, incorporating the prior knowledge of the specific downstream task into the pretraining procedure may be extremely valuable. Frangi filtering is an effective technique to enhance and extract tubular structures in medical images \cite{frangi1998multiscale}. The filtering process involves analyzing the local intensity and Hessian matrix of the image to detect vessel-like structures, making it useful for applications such as vessel segmentation and analysis in medical imaging. Frangi filtering plays a fundamental role as a preprocessing step in traditional vessel segmentation algorithms. It is worth seriously considering incorporating this filtering technique into the pretraining task.

Driven by the abovementioned analysis, we propose a Frangi filtering-based pretraining network (called FFPN) to fully use the public unlabeled TOF-MRA data. In particular, we have developed a preprocessing workflow to handle large-scale unlabeled data. This preprocessing program utilizes Frangi filtering to enhance vessel structures. Additionally, it incorporates thresholding methods and connected component analysis to obtain a coarse segmentation of the vessels. Then, the vessel-enhanced images and coarse vessel segmentations are used for a multi-task pretraining procedure. Our model is pretrained on the large-scale TOF-MRA dataset with 1510 volumetric data and evaluated on four labeled datasets. The results demonstrate that it notably outperforms the state-of-the-art SemiSL and SSL methods. Besides, the ablation studies also confirm that our proposed pretraining task helps the model achieve better performance with fewer labeled data. Moreover, this pretraining task is not limited to specific model architectures and can significantly improve the performance across various backbone structures consistently.

The main contributions of this study can be summarized as follows: 
\begin{itemize}
    \item We develop an automated preprocessing workflow to efficiently handle unlabeled data. The preprocessing provides the vessel enhanced images and coarse vessel segmentations that are used for the pretraining procedure. 
    \item We propose a simple yet effective pretraining strategy by using the preprocessed TOF-MRA data. This pretraining task contains regression learning, coarse label segmentation, and max intensity projection (MIP) consistency learning. The results show that the proposed method shows great superiority over existing models.
    \item We constructed a large unlabeled dataset (1510 volumetric data) and three manual-annotated datasets (a total of 113 volumetric data). These data are expected to advance semi-supervised, weakly supervised, and self-supervised learning methods in cerebrovascular segmentation.
\end{itemize}
\begin{figure*}[t]
\centering
\includegraphics[width=\textwidth]{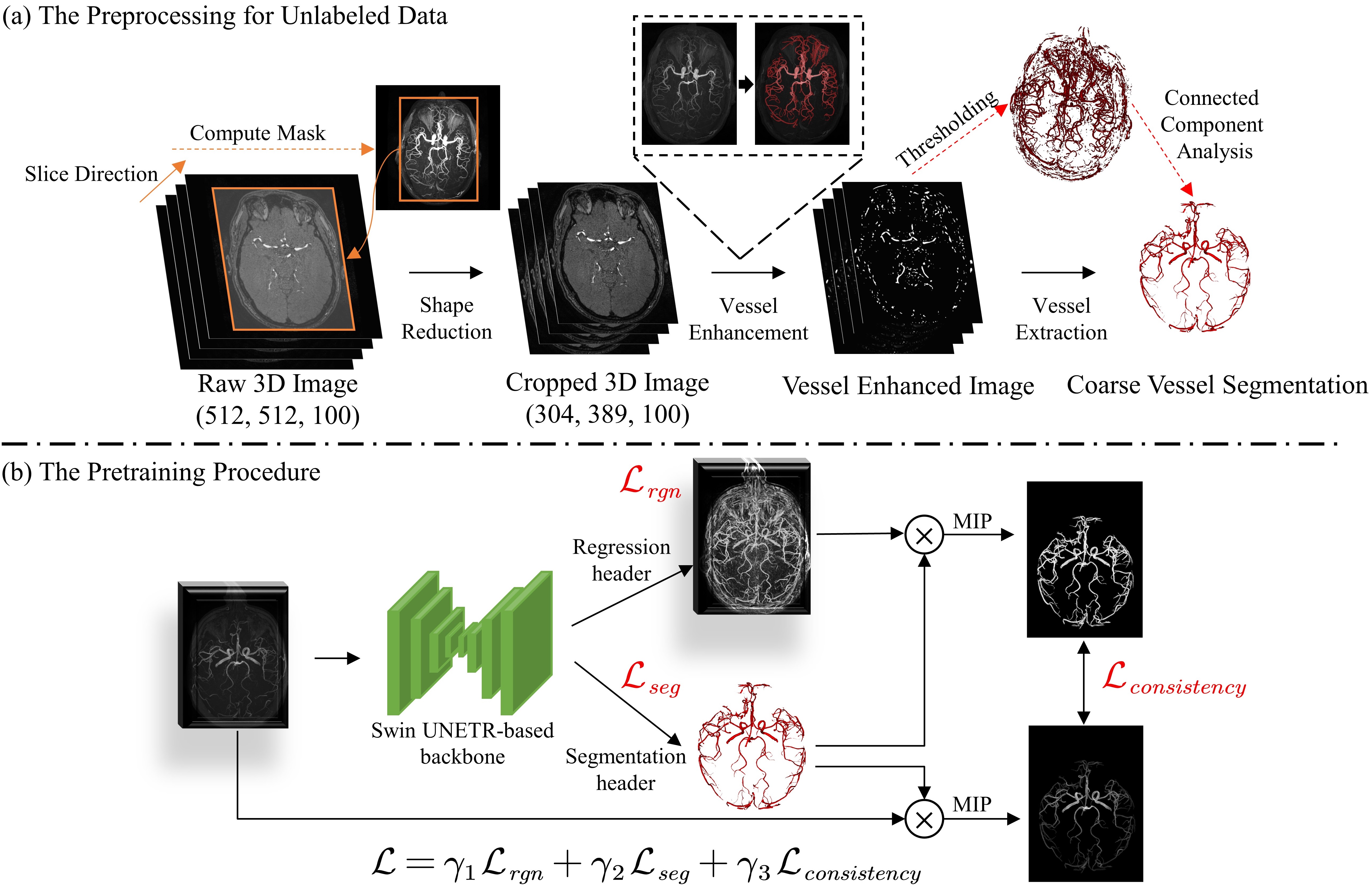}
\caption{(a) The illustration of our proposed preprocessing workflow. (b) The overall framework of our proposed pretraining procedure with multi-task learning.}
\label{fig:framework}
\end{figure*}

\section{Related Work}
\subsection{Cerebrovascular Segmentation}
Traditional methods for cerebrovascular segmentation typically rely on the continuity of grayscale variations in images to perform threshold-based segmentation \cite{otsu1979threshold, frangi1998multiscale}. In addition, statistical modeling-based approaches utilize the concept that different brain tissues exhibit distinct pixel grayscale distributions during imaging \cite{duan2019segmentation, alderliesten2006towards}. This has been successful in the segmentation of cerebrovascular structures in TOF-MRA images. In recent years, deep learning methods have shown remarkable performance in vessel segmentation tasks \cite{sanchesa2019cerebrovascular, sichtermann2019deep, qi2023dynamic}. For instance, Livne et al. utilized a UNet-based model to segment vessels in patients with cerebrovascular disease \cite{livne2019u}. Chen et al. proposed an attention-assisted generative adversarial network (A-SegAN) for automatic cerebrovascular segmentation \cite{chen2022attention}. More related research can be referred to this review \cite{chen2023all}.

\subsection{SemiSL and SSL Methods for Medical Images}
Numerous SemiSL and SSL methods have been developed to utilize unlabeled medical data and improve model performance \cite{wang2023deep, YU2023103062, 10189221, liao2023cupre}. For instance, Chen et al. proposed a generative consistency for the semi-supervised (GCS) model, which calculates the consistency of perturbed data to enhance cerebrovascular segmentation performance \cite{chen2022generative}. Moreover, SSL-based methods have also demonstrated the effectiveness of pretraining in medical imaging research. To illustrate, Tang et al. introduced a multitask pretraining strategy using a 3D Swin transformer, achieving significant success in multiple organ segmentation tasks. For more detailed information on the topic, comprehensive reviews \cite{zhang2023dive, jiao2022learning, krishnan2022self} can be referred to.

\section{Methods}
In this section, we introduce the proposed FFPN model. The preprocessing workflow and pretraining procedure are shown in Fig.\ref{fig:framework}~a and b, respectively. First, we preprocess the unlabeled TOF-MRA data to obtain the vessel enhanced image (VEI) and coarse vessel segmentation (CVS). Subsequently, during the pretraining procedure, the VEI and CVS are utilized as regression and segmentation targets. Further details regarding this process will be elaborated in the upcoming sections.  

\subsection{Preprocessing Method for Unlabeled Data}
\label{preprocessing workflow}
The pretraining process consists of three steps: image size cropping, vessel-structure enhancement, and coarse segmentation.

\subsubsection{Image size cropping:}
The TOF-MRA images often contain a significant number of background pixels, which have no relevance to the vessel segmentation task. Removing these background pixels provides a key advantage by reducing computational costs, especially when dealing with a large-scale unlabeled dataset. Previous approaches have utilized non-zero masks to crop raw images \cite{isensee2021nnu}. However, background pixels don’t always possess intensity values of zero. In this study, we propose a novel method to calculate the cropping mask based on the variation of pixel intensities along the Z-axis (slice direction).

For a given single image, denoted as $I_i$ with dimensions $(H, W, D)$, we compute three projection images along the Z-axis: Average Intensity Projection (AIP), Maximum Intensity Projection (MIP), and Intensity Variation Map (IVM). The calculations are as follows:

\begin{align}
&AIP_i = \mathrm{Average}(I_i) \\ \notag
&MIP_i = \mathrm{Maximum}(I_i) \\ \notag
&IVM_i = \mathrm{STD}(I_i)
\end{align}

To create the cropping mask, we employ threshold segmentation. Taking MIP as an example, we generate a mask by retaining only the positions of the top percentage (e.g., 35\%) of pixels in the MIP image. The resulting masks, namely the AIP mask, MIP mask, and IVM mask, are then merged using a bitwise AND operation. Additionally, small regions containing fewer than 200 pixels are removed. Finally, we obtain a square cropping mask that is applied to each slice of $I_i$.

In summary, our proposed approach enables the generation of a cropping mask based on the intensity variations along the Z-axis in TOF-MRA images. By applying this mask, background pixels are effectively eliminated, resulting in a significant reduction in computational costs.
\subsubsection{Frangi fitering:}
For 3D TOF-MRA images, the Frangi filtering involves several steps to enhance the visibility of blood vessels. First, it computes the second-order mixed partial derivatives for each pixel using the Hessian matrix. The Hessian matrix is defined as:
\begin{equation}
H = 
\begin{bmatrix}
\frac{\partial^2 I}{\partial x^2} & \frac{\partial^2 I}{\partial x \partial y} & \frac{\partial^2 I}{\partial x \partial z} \\
\frac{\partial^2 I}{\partial y \partial x} & \frac{\partial^2 I}{\partial y^2} & \frac{\partial^2 I}{\partial y \partial z} \\
\frac{\partial^2 I}{\partial z \partial x} & \frac{\partial^2 I}{\partial z \partial y} & \frac{\partial^2 I}{\partial z^2} \\
\end{bmatrix}
\end{equation}
where $\frac{\partial^2 I}{\partial x^2}$ represents the second partial derivative of the image intensity $I$ with respect to the $x$ coordinate, and similarly for the other partial derivatives. Next, the eigenvalues of the Hessian matrix, denoted as $\lambda_1$, $\lambda_2$, and $\lambda_3$, are calculated. These eigenvalues provide information about the local structure of the image. Specifically, the magnitude and orientation of the eigenvalues help identify pixels that belong to blood vessels.

In general, larger positive eigenvalues correspond to the central region of vessels, smaller positive eigenvalues represent the vessel edges, while negative eigenvalues typically indicate irrelevant structures and backgrounds. By analyzing the eigenvalues, pixels with vascular features are selected, and the VEI is obtained at this step.
\subsubsection{Coarse segmentation:}
To obtain a coarse segmentation of the blood vessels, a thresholding method and connected component analysis are applied. First, a vessel mask is created by retaining only the positions of the top percentage (e.g., 5\%) of pixels for VEIs. This thresholding step helps separate the vessel pixels from the background and other structures. Then, connected component analysis is performed on the vessel mask. The largest $k$ connected regions, where $k$ could be empirically chosen between 3-5, are retained as the coarse segmentation result of blood vessels. By combining the Frangi filtering, thresholding, and connected component analysis, the CVS is obtained, which provides an initial approximation of the blood vessel structure in the 3D TOF-MRA images.
\begin{table*}[t]
\centering
\fontsize{9pt}{10pt}\selectfont
\begin{tabular}{lllllllllllll}
\hline
Datasets                & \multicolumn{3}{c}{TubeTK-42}                          & \multicolumn{3}{c}{Brains}                          & \multicolumn{3}{c}{IXI-45}                             & \multicolumn{3}{c}{EDEN}                            \\
Method    & Dice             & clDice           & HD95          & Dice             & clDice           & HD95          & Dice             & clDice           & HD95          & Dice             & clDice           & HD95          \\\hline
UNet (MICCAI'15)      & 72.98             & 81.33            & 9.09           & 73.77             & 75.86             & 3.72              & 79.82             & 75.23             & 10.25             & 79.79          & 80.62          & 9.89          \\\hline
DTC (AAAI'21)       & 69.31             & 76.35            & 25.60          & 64.36             & 64.86             & 34.72             & 70.44             & 63.48             & 63.31             & 71.65          & 65.65          & 91.93         \\
GCS (TMI'22)       & 18.11             & 33.40            & 22.50          & 10.37             & 9.28              & 24.92             & 25.05             & 21.32             & 55.10             & 61.36          & 67.85          & 28.42         \\
GBDL (CVPR'22)      & 36.40             & 55.67            & 16.16          & 39.79             & 52.84             & 63.71             & 12.46             & 20.37             & 42.43             & 51.78          & 64.48          & 11.41         \\
SSNet (MICCAI'22)     & 65.15             & 71.84            & 28.65          & 64.22             & 63.88             & 35.04             & 73.55             & 65.14             & 52.05             & 69.72          & 64.26          & 96.65         \\
BCP (CVPR'23)       & 73.36             & 81.95            & 8.33           & 77.25             & 80.31             & 2.88              & 81.37             & 80.64             & 6.17              & 86.44          & \underline{89.41}          & 2.71          \\\hline
PCRL (ICCV'21)      & 72.90             & 81.92            & \underline{7.82}& 75.94            & 79.09             & \underline{2.74}  & 83.16             & 83.69             & 4.05              & 86.15          & 88.71          & 2.91          \\
MAE (CVPR'22)       & 72.39             & 79.37            & 11.82          & 77.56             & 79.70             & 3.49              & 83.21             & 80.14             & 10.04             & 84.62          & 85.90          & 6.73          \\
Swin UNETR (CVPR'22)& \underline{74.66} & 82.18            & 9.71           & 77.76             & \underline{80.40} & 3.30              & 80.64             & 79.01             & 6.37              & 81.23          & 81.87          & 12.90         \\
UniMiSS (ECCV'22)   & 70.54             & \underline{82.59}& 9.03           & 77.63             & 79.62             & 3.03              & 85.45             & 83.49             & 3.83              & \underline{88.02}          & 88.51          & 2.91          \\
GVSL (CVPR'23)      & 74.27             & 82.57            & 8.30           & \underline{77.87} & 79.68             & 3.16              & \underline{85.92} & \underline{84.82} & \underline{3.35}  & 85.15          & 89.39          & \underline{2.61}          \\\hline
FFPN (Ours)      & \textbf{77.53} & \textbf{85.71} & \textbf{6.69} & \textbf{80.47} & \textbf{83.28} & \textbf{2.19} & \textbf{87.24} & \textbf{86.25} & \textbf{2.75} & \textbf{90.75} & \textbf{91.53} & \textbf{1.24} \\
Preprocessing    & 29.64          & 31.27          & 71.29          & 31.73             & 37.01            & 38.55    & 43.18             & 55.50             & 38.56             & 68.48          & 69.43          & 46.18   \\ \hline
\end{tabular}
\caption{The cerebrovascular segmentation results on TubeTK-42, Brains, IXI-45 and EDEN datasets}
\label{tab:seg_results}
\end{table*}

\subsection{Pretraining task and Loss Function}
\label{pretraining tasks}
Our proposed pretraining process contains three tasks: regression learning, coarse vessel segmentation learning, and consistency learning in the MIP projection. Let function $f(\cdot)$ be the backbone neural network, and $V_i$ and $S_i$ are the VEI and CVS for subject $i$. $f(\cdot)$ is used to encode the raw images, and one regression header $Rgn(\cdot)$and $Seg(\cdot)$segmentation header are used to predict $V_i$ and $S_i$, respectively. It can be formulated as:

\begin{align}
&\hat{V}_i = Rgn(f(I_i)) \\ \notag
&\hat{S}_i = Seg(f(I_i)) \\ \notag
\end{align}
where $Rgn(\cdot)$and $Seg(\cdot)$ are convolution operation with kernel size 1. $\ell_1$ is used to compute regression loss. For segmentation loss, Dice loss and Binary Cross Entropy are used. These can be formulated as:
\begin{equation}
    \mathcal{L}_{rgn}= \|\hat{V}_i - V_i\|_1
\end{equation}
\begin{equation}
\mathcal{L}_{seg}= \mathrm{Diceloss}(\hat{S}_i - S_i) + \mathrm{BCEloss}(\hat{S}_i - S_i)
\end{equation}

In this study, we do not conduct skull stripping for the preprocessed data. In such case, the Frangi filtering will also enhance the skull pixels, while the CVS will not contain the skull pixels after connected component analysis. In order to focus more on learning the vascular structures, we incorporate an extra consistency loss by calculating the $\ell_1$ loss between $\hat{S}_i \odot \hat{V}_i$ and $\hat{S}_i \odot I_i$ after applying maximum intensity projection along Z-axis. It can be formulated as:
\begin{align}
& \mathrm{MIP}_{VS} = \mathrm{Maximum}(\hat{S}_i \odot \hat{V}_i) \\ \notag
& \mathrm{MIP}_{IS} = \mathrm{Maximum}(\hat{S}_i \odot \hat{I}_i) \\ \notag
& \mathcal{L}_{consistency} = \|\mathrm{MIP}_{VS} - \mathrm{MIP}_{IS} \|_1
\end{align}
where $\hat{S}_i \odot \hat{V}_i$ and $\hat{S}_i \odot I_i$ have almost removed the skull pixels. Meanwhile, the MIP images of TOF-MRA are commonly used for the initial assessment of vascular morphology. The consistency loss will help the model focus more on learning the vascular structures.

Finally, the total loss function $\mathcal{L}$ will be:
\begin{equation}
    \mathcal{L} = \gamma_1\mathcal{L}_{rgn} + \gamma_2\mathcal{L}_{seg} + \gamma_3\mathcal{L}_{consistency}
\end{equation}
where $\gamma_1, \gamma_2$ and $\gamma_3$ are the loss weight. We empirically set $\gamma_1=0.4, \gamma_2=0.4$ and $\gamma_3=0.2$ in this study.
\begin{figure}
\centering
\includegraphics[width=\linewidth]{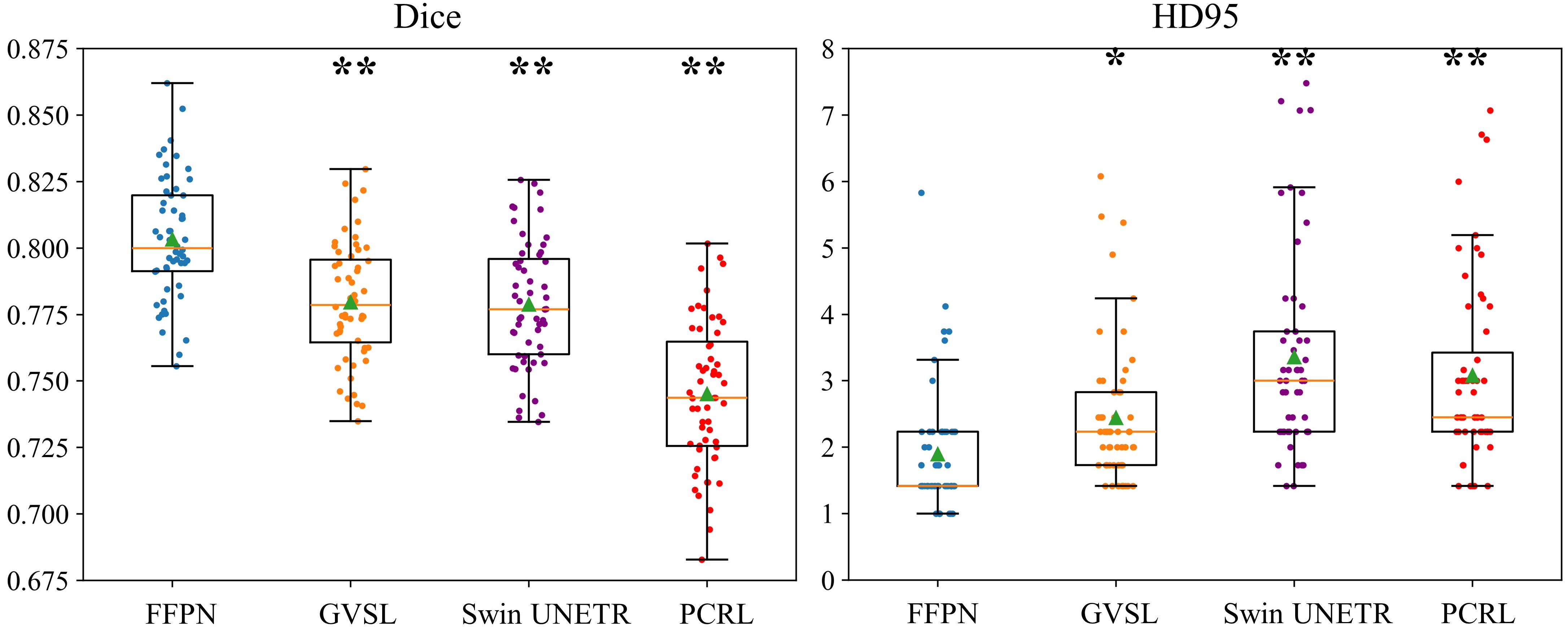}
\caption{Statistical validation on Brains dataset by using the T-test. '*' indicates $p<0.05$ and '**' indicates $p<0.001$.}
\label{fig:statistical result}
\end{figure}

\subsection{Backbone}
We select the Swin UNETR \cite{hatamizadeh2021swin} as the backbone in this study. Different from the vanilla model, successive convolutional layers are used in the initial embedding layers \cite{zhou2021nnformer}. This setting enhances the capture of spatial location information, a feature that is potentially beneficial to the segmentation of cerebrovascular structures with sparse distribution. Please note that we also examined the impact of our pretraining strategy on other backbone structures in the result section.

\begin{figure*}
\centering
\includegraphics[width=\textwidth]{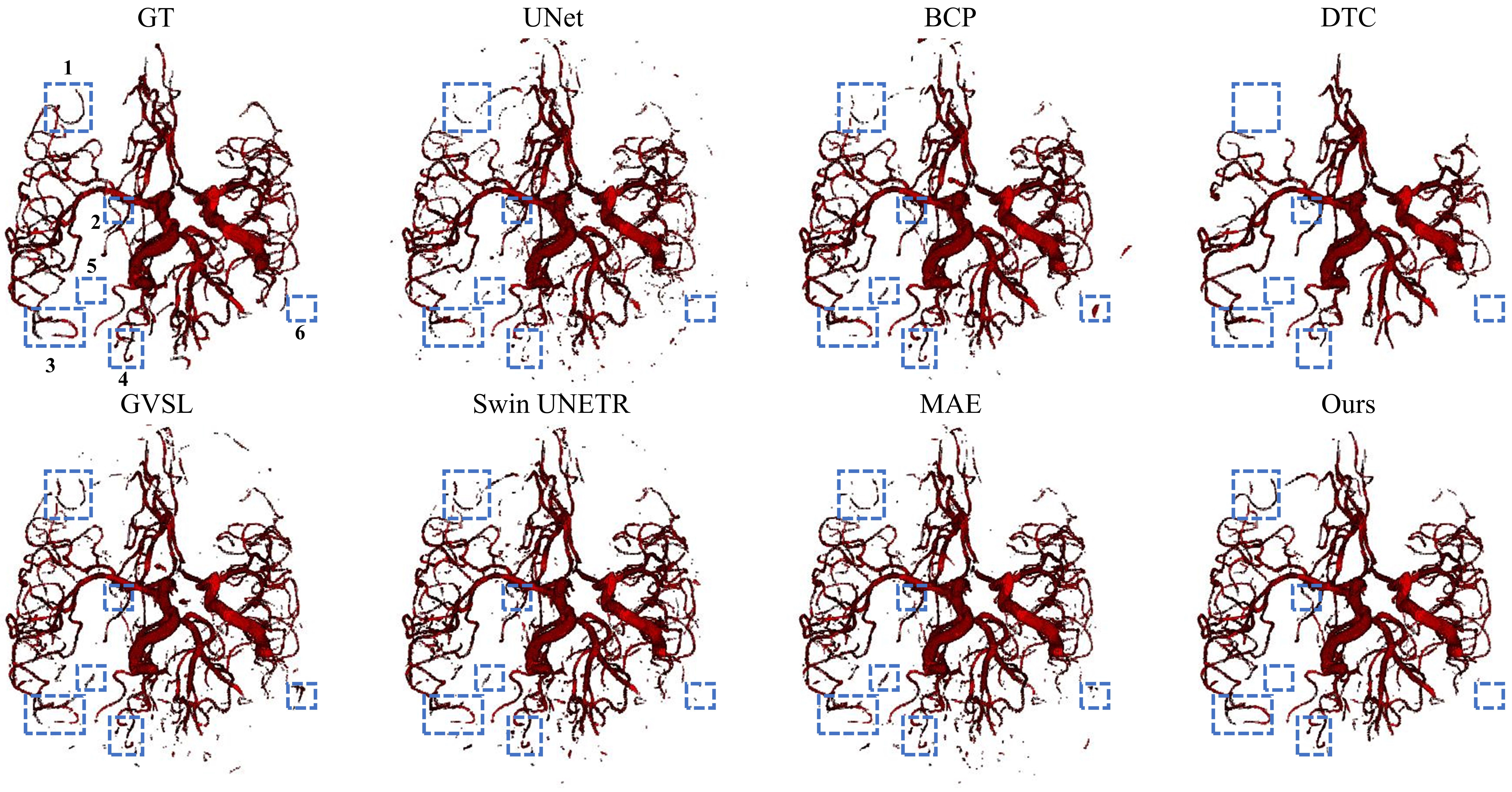}
\caption{The visualization comparison results of our method and the competing baseline models from the 3D view.}
\label{fig:3D_VIEW}
\end{figure*}

\section{Experimental Setup}
 \subsection{Datasets}
\noindent\textbf{Unlabeled Datasets for Pretraining:} We collect a large unlabeled TOF-MRA consisting of 1510 subjects from 5 public datasets, including IXI (525 subjects), OASIS3 (525 subjects) \cite{lamontagne2019oasis}, BrainAneurysm (280 subjects) \cite{di2023towards}, ADAM (113 subjects) and TubeTK (67 subjects). We preprocessed these data by our proposed workflow and the model is pretrained through the proposed tasks.

\noindent\textbf{Labeled Datasets for Downstream Evaluation:} We evaluate our proposed method in four labeled datasets.
\begin{itemize}
    \item \textbf{TubeTK-42}: The full TubeTK dataset contains 109 subjects, with 67 unlabeled subjects used for pretraining and 42 labeled subjects used for finetune validation. The image size is (448, 448, 128) and the spacing is (0.5134, 0.5134, 0.8000).
    \item \textbf{Brains}: This dataset contains 56 subjects. The size of the images is not identical, with average size of (286, 320, 204). The spacing is (0.6188, 0.6188, 0.6200).
    \item \textbf{IXI-45}: This dataset is provided by \cite{chen2022attention}. It contains 45 subjects from the whole IXI dataset (570 subjects). The image size is (1024, 1024, 92). The provided images have been processed and therefore do not contain spacing information. The original spacing of the IXI dataset is (0.2637, 0.2637, 0.8000).
    \item \textbf{EDEN}: This dataset contains 15 subjects. Unlike the previous dataset, we also provide the blood vessels on the skull in this dataset. The image size is (672, 672, 210) and the spacing is (0.2976, 0.2976, 0.4500).
\end{itemize}
Please note that the pretraining dataset and evaluation dataset are completely isolated. For the labeled datasets, 80\% subjects are used for training, and the rest data are used for testing. Detailed information about the dataset can be found in the Appendix.

\subsection{Evaluation Metrics}
Three metrics are used to assess the cerebrovascular segmentation performance, including the Dice similarity coefficient (Dice), Hausdorff Distance 95\% (HD95) and centerline Dice coefficient (clDice) \cite{cldice2021}. The Dice and HD95 metrics primarily focus on the overlap degree and shape differences in segmentation results, clDice takes into account the topological structure of the segmented regions, making it more suitable for evaluating tubular structures.

\subsection{Baseline Models and Implementation}
We compare the performance of 3D cerebrovascular segmentation between the proposed FFPN model and existing methods using unlabeled data, including the SemiSL methods BCP \cite{bai2023bidirectional}, GBDL \cite{wang2022rethinking}, GCS \cite{chen2022generative}, SS-Net \cite{wu2022exploring}, DTC \cite{luo2021semi} and the SSL methods MAE \cite{he2022masked}, PCRL \cite{zhou2021preservational}, UniMiSS \cite{xie2022unimiss}, Swin UNETR \cite{tang2022self}, GVSL \cite{he2023geometric}. Besides, we also include the basic UNet \cite{ronneberger2015u} model that does not utilize the unlabeled data to provide a comparison.

For our proposed FFPN model, the encoder has 4 stages which comprise of [2, 2, 6, 2] transformer blocks at each stage. The initial patch size is 2 and the feature size is 24. In pretraining task learning, the model is trained using AdamW optimizer for 200k iterations. Warm-up cosine scheduler of 20k iterations is used. The cropping patch size is (160, 160, 64) and the batch size is 2. The weight decay is $3\times10^{-5}$ and learning rate is $5\times10^{-3}$. In finetune procedure, the model is trained using AdamW optimizer for 5k iterations. The learning rate is $5\times10^{-3}$ (half the learning rate for the encoder) and the cropping patch size is (192, 192, 96).

\section{Experimental Results}
\subsection{Comparison with state-of-the-art Methods}
The cerebrovascular segmentation results for the four datasets are shown in Table \ref{tab:seg_results}. The proposed method achieves the best performance compared with other baseline models on the four datasets in terms of the three metrics. In TubeTK-42 and Brains datasets, the clDice scores of our proposed method are 85.71 and 83.28, with an improvement of approximately 3\% over the second model. In IXI-45 and EDEN datasets, our proposed method also achieves an improvement of around 2\% compared with the second-best model. 

To further demonstrate the superiority of FFPN, a statistical analysis is shown in Fig. \ref{fig:statistical result}. Specifically, We conduct another 5-fold experiment on the largest dataset (Brains) using FFPN and baseline models with competitive performance (GVSL, Swin UNETR and PCRL). The results show that the improvement of FFPN is statistically significant (T-test) among the three metrics (the outcome for the metric clDice aligning closely with that of Dice).

We provide a 3D visualization result to support an intuitive evaluation (see Fig. \ref{fig:3D_VIEW}). Compared with other methods, the segmentation produced by FFPN reveals marked superiority in achieving continuity for the vessel structure, particularly noticeable at locations 1, 2, 3 and 4. Furthermore, it is worth highlighting that the FFPN-enabled segmentation leads to a reduction in false positive points (e.g., locations 5 and 6).

\begin{figure}
\centering
\includegraphics[width=\linewidth]{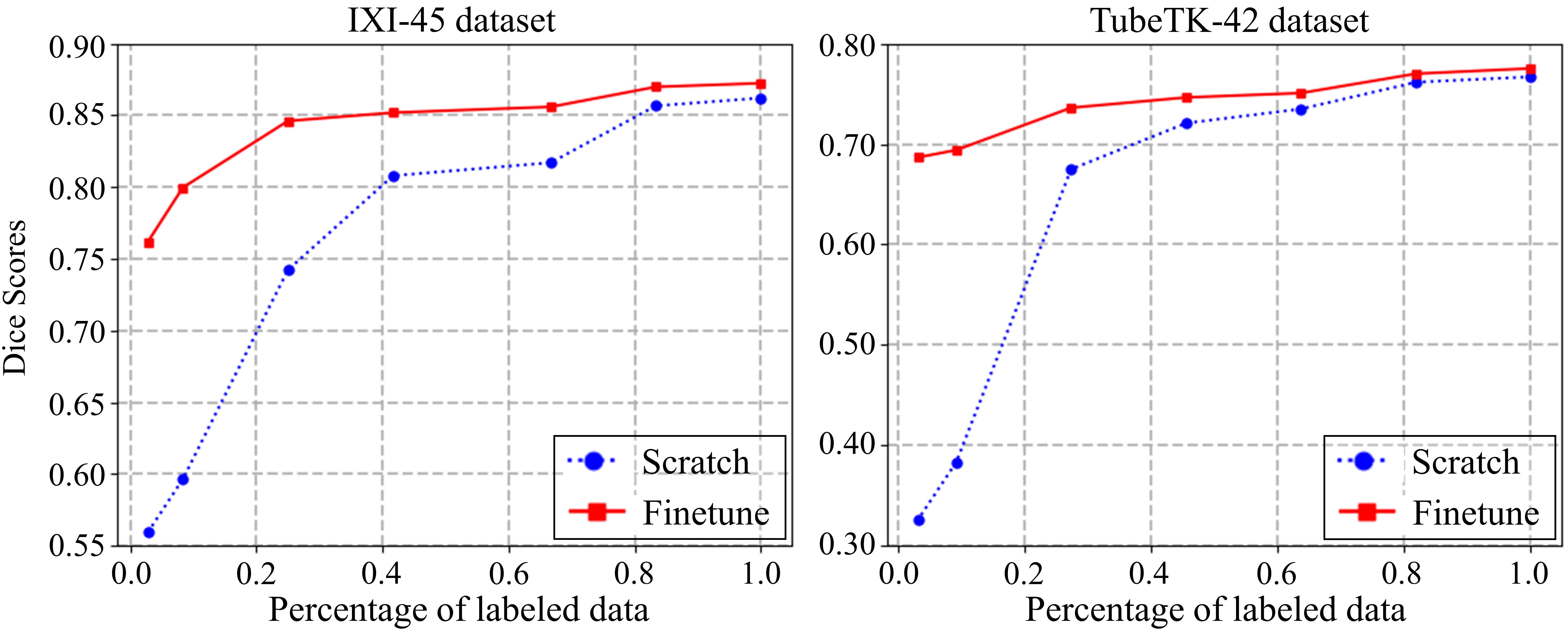}
\caption{Variation of the effect of the pretraining strategy with the labeled data percentage.}
\label{fig:labeled_number}
\end{figure}

\begin{table}
\centering
\begin{tabular}{llll}
\hline
Loss Function                   & Dice   & clDice & HD95    \\ \hline
Scratch                         & 72.08 & 80.16 & 11.46 \\ \hline
$\mathcal{L}_{rgn}$             & 72.34 & 81.11 & 9.62  \\
$\mathcal{L}_{seg}$             & 73.46 & 81.58 & 9.10  \\
$\mathcal{L}_{consistency}$     & 72.15 & 80.78 & 10.13        \\
$\mathcal{L}_{rgn}+\mathcal{L}_{seg}$ & 74.18 & 82.17 & 8.14  \\
$\mathcal{L}_{rgn}+\mathcal{L}_{seg}+\mathcal{L}_{consistency}$ & \textbf{74.64} & \textbf{82.94} & \textbf{7.57}  \\ \hline
\end{tabular}
\caption{The effectiveness of different loss functions.}
\label{tab:loss_func}
\end{table}

\subsection{Influence of the Percentage of Manual Labeled Data}
We then conduct a study to evaluate the effectiveness of our pretraining method using a reduced amount of manually labeled data. The results are presented in Fig. \ref{fig:labeled_number}.

It is observed that pretraining consistently improves the performance of the model across different proportions of labeled data. When using all available labeled data, our pretraining method achieves a 1.03\% higher Dice score for the IXI-45 dataset and 0.83\% higher Dice score for the TubeTK-42 dataset compared to a model trained from scratch. Additionally, with less than half of the labeled data, the pretrained model demonstrates an improvement of 4.39\% for the IXI-45 dataset and 2.56\% for the TubeTK-42 dataset. Furthermore, as the availability of labeled images decreases, the performance gap between the models trained with and without pretraining increases.

\subsection{Effectiveness of Pretraining Loss Function}
In this section, we evaluate the effectiveness of the proposed three loss functions. Specifically, The model is pretrained using different loss functions and then their effectiveness is assessed during the finetuning procedure. We choose the TubeTK-42 dataset with 45.45\% labeled data (i.e., 15 subjects) as an example, and the results are presented in Table \ref{tab:loss_func}.

For single-task pretraining, the segmentation learning achieves the best performance with metrics Dice, clDice, and HD95 values of 73.46, 81.58, and 9.10, respectively. The regression learning also shows similar performance, while the consistency learning does not significantly improve the model’s performance. Moreover, when combining the regression and segmentation learning loss functions, the model’s performance is further improved. Taking into account all three learning tasks, the model attains the highest performance.

\begin{figure}
\centering
\includegraphics[width=\linewidth]{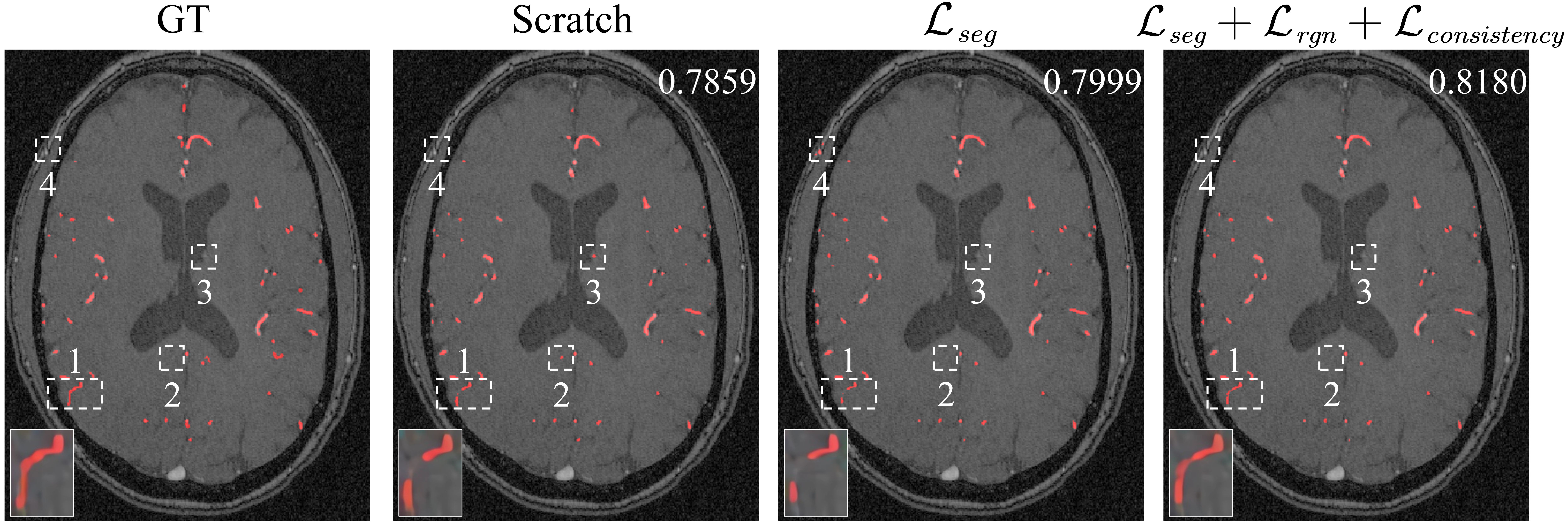}
\caption{Illustration of the effectiveness of pretraining loss functions from the slice view. The number indicates the clDice metric.}
\label{fig:visu_loss_func}
\end{figure}

We also present a visualization result from the slice view in Fig. \ref{fig:visu_loss_func}. Incorporating pretraining with $\mathcal{L}_{seg}$ aids in reducing the occurrence of false positive points in the generated output (e.g., locations 2 and 3). However, it is worth noting that due to the coarse segmentation results potentially retaining some information from the skull, there may still be instances where certain structures on the skull are annotated (e.g., location 4). Overall, by incorporating all pretraining tasks, the model achieves improved segmentation, characterized by enhanced continuity of vessels (see location 1) and a reduced presence of false positive points.

\begin{table}
\centering
\begin{tabular}{llll}
\hline
Loaded component  & Dice            & clDice        & HD95              \\\hline
Scratch           & 80.79          & 78.08          & 8.41        \\\hline
Encoder           & 81.71          & 79.90          & 12.08          \\
Decoder           & 81.14          & 76.22          & 11.19           \\
Encoder + Decoder & 83.77          & 81.57          & 6.59          \\
All               & \textbf{85.19} & \textbf{83.09} & \textbf{4.43} \\ \hline
\end{tabular}
\caption{The effectiveness of different model component pretraining parameters loaded.}
\label{tab: model_component}
\end{table}

\subsection{Influence of Model Components for Pretraining}
Different from previous SSL methods that typically only load the encoder parameters for downstream tasks, our proposed approach loads both the encoder and decoder parameters, as well as the segmentation header. In this section, we explore the impact of loading different model components. We take the IXI-45 dataset with 41.67\% labeled data (i.e., 15 subjects) as an example, and the results are presented in Table \ref{tab: model_component}.

When only the encoder is loaded, there is a notable improvement in performance, with the clDice score increasing to 79.90. Besides, when only the decoder is loaded, there is a drop in performance compared to the encoder-loaded model, as indicated by the clDice score of 76.22. Furthermore, when both the encoder and decoder are loaded, there is a substantial boost in performance. The clDice score increases to 81.57. When all components, including the segmentation header, we observe the highest performance across all metrics. 

\begin{table}
\centering
\begin{tabular}{llll}
\hline
Dataset       & \multicolumn{3}{c}{EDEN}         \\
Backbone      & Scratch & Finetune & Improvement \\ \hline
UNet          & 79.79  & 84.26   & +4.47\%      \\
AttentionUNet & 78.44  & 88.20   & +9.76\%      \\
UNETR         & 83.32  & 88.06   & +4.74\%      \\
Swin UNETR    & 88.87  & 90.75   & +1.88\%      \\\hline
Dataset       & \multicolumn{3}{c}{IXI-45}          \\
Backbone      & Scratch & Finetune & Improvement \\\hline
UNet          & 79.82  & 85.24   & +5.41\%      \\
AttentionUNet & 81.52  & 84.62   & +3.10\%      \\
UNETR         & 80.01  & 83.85   & +3.84\%      \\
Swin UNETR    & 86.21  & 87.24   & +1.03\%     \\ \hline
\end{tabular}
\caption{The effectiveness of our proposed pretraining method on various backbone structures. The number indicates the Dice scores.}
\label{tab: backbone}
\end{table}

\subsection{The Effectiveness on Different Backbone Structures}
\label{section:backbone}
In this section, we assess the effectiveness of our proposed pretraining learning on different backbone structures. We focus on four fundamental medical segmentation backbones, which encompass both convolutional neural network (CNN)-based models (UNet \cite{ronneberger2015u} and AttentionUNet \cite{oktay2018attention}) and Transformer-based models (UNETR \cite{hatamizadeh2022unetr} and Swin UNETR \cite{hatamizadeh2021swin}). The corresponding results are summarized in Table \ref{tab: backbone}.

For the EDEN dataset, we observe improvements when using our proposed pretraining learning compared to training from scratch across all backbone architectures. Specifically, for the UNet backbone, the Dice score increases from 79.79 to 84.26, resulting in a significant improvement of 4.47\%. Similarly, the AttentionUNet and UNETR backbones show improvements of 9.76\% and 4.74\%, respectively. Even for the advanced Swin UNETR backbone, there is still a notable 1.88\% improvement.

In the case of the IXI-45 dataset, we observe similar trends. Our proposed pretraining learning consistently leads to improvements compared to training from scratch. The UNet backbone shows an improvement of 5.41\%, while the AttentionUNet, UNETR, and Swin UNETR backbones show improvements of 3.10\%, 3.84\%, and 1.03\%, respectively.

\subsection{The Robustness of Our Pretraining Method Regarding the Unlabeled Dataset Heterogeneity}
We carried out this experiment to assess the robustness of our proposed method in relation to multi-source heterogeneous pretraining data, using the IXI dataset as an illustrative case.

\begin{figure}
\centering
\includegraphics[width=\linewidth]{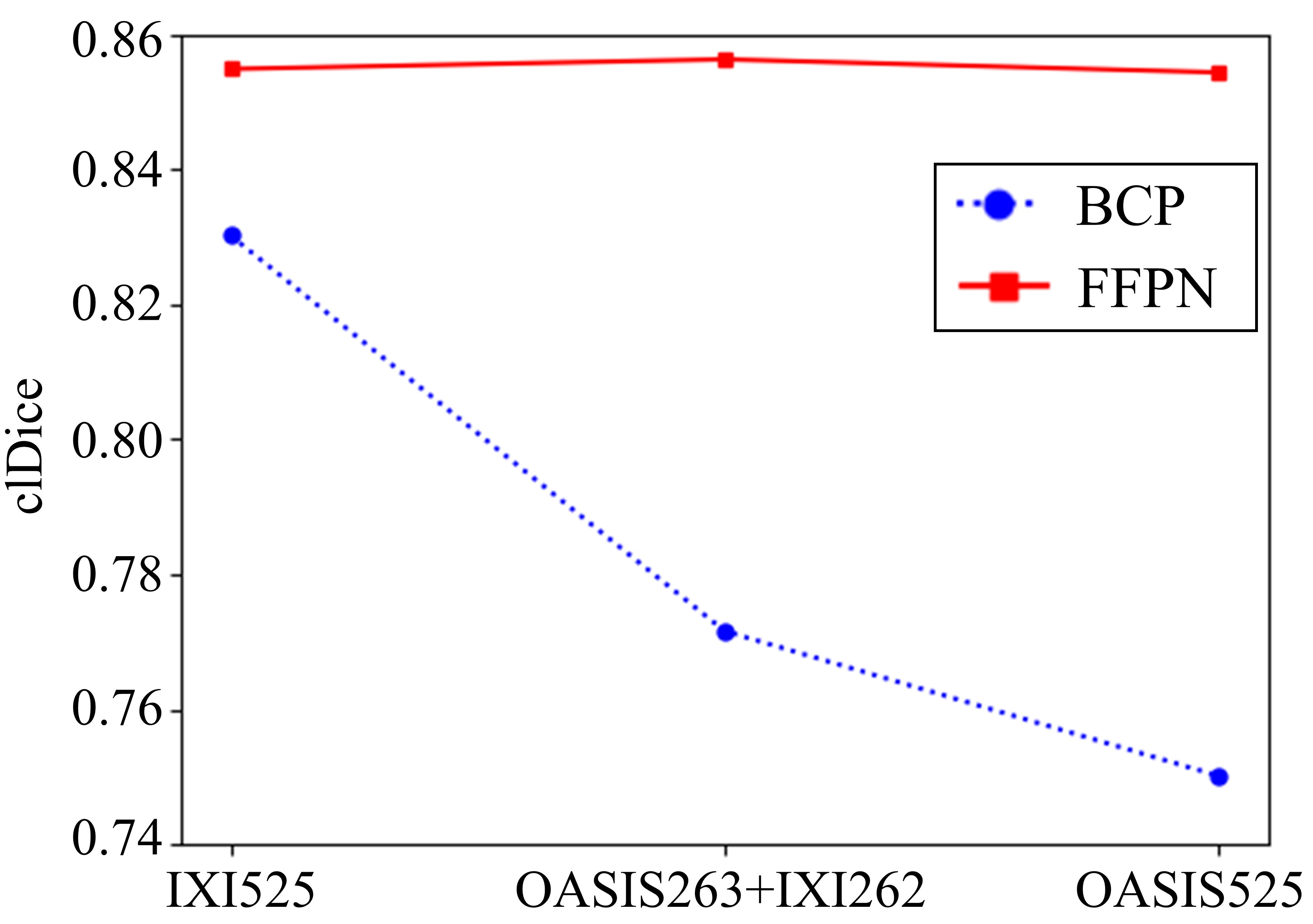}
\caption{The robustness of our proposed method. Horizontal coordinates indicate the composition of unlabeled data..}
\label{fig:robustness}
\end{figure}

In particular, we maintain the quantity of unlabeled data constant (525 subjects) while varying the proportion of the unlabeled IXI dataset used. Then we test the model performance on the labeled IXI-45 dataset. The best-performing semiSL method BCP is used for comparison and The outcomes are depicted in Figure \ref{fig:robustness}. From these results, it is observed that the performance of the semi-supervised learning method, BCP, significantly diminishes as the number of homogeneous data decreases. Conversely, our proposed method consistently shows a stable and high segmentation performance. This result demonstrates that the proposed pretraining strategy may maintain a high level of robustness.

\section{Discussion}
To unleash the potential of unlabeled TOF-MRA data for 3D cerebrovascular segmentation, our approach incorporates Frangi filtering as a key element in the pretraining phase. The proposed FFPN  model demonstrates enhanced robustness and superior segmentation performance, notably outshining other SemiSL and SSL methodologies.

\begin{figure}
\centering
\includegraphics[width=\linewidth]{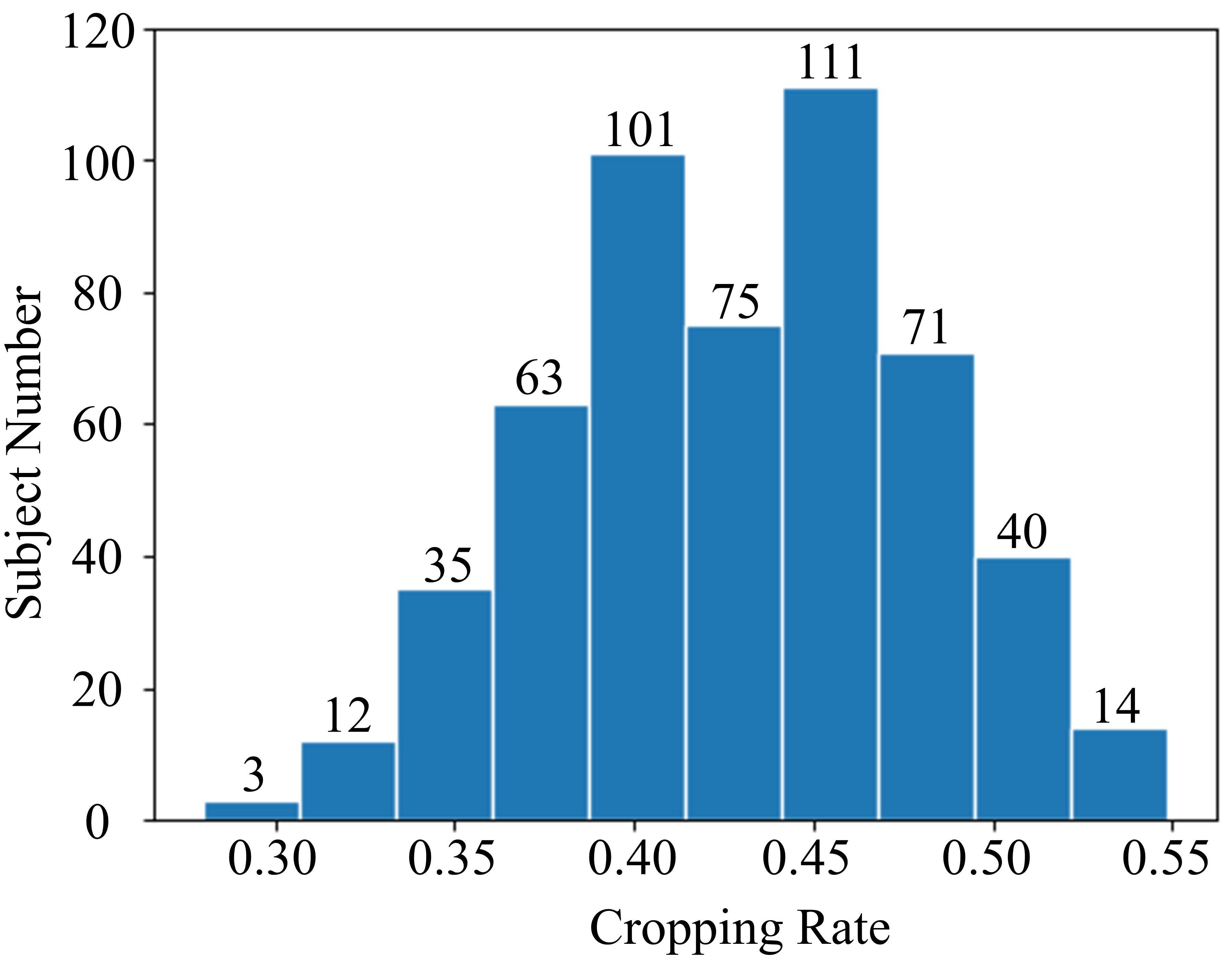}
\caption{The cropping rate in IXI dataset.}
\label{fig:cropping effectiveness}
\end{figure}

\begin{figure}
\centering
\includegraphics[width=\linewidth]{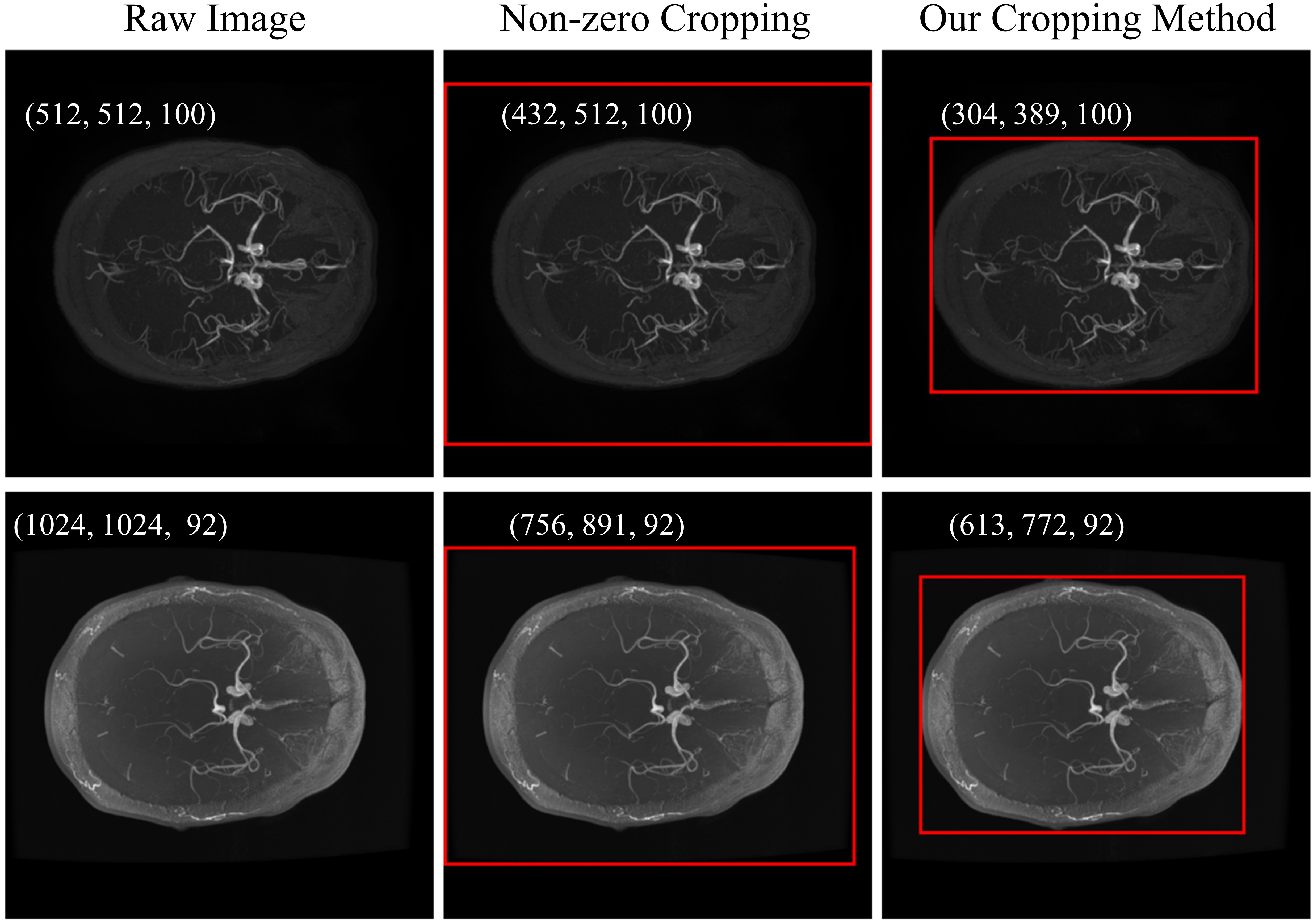}
\caption{Three representative subjects for the cropping strategy. The number indicates the image size.}
\label{fig:cropping view}
\end{figure}

Our proposed cropping strategy can reduce the computation cost when processing such a large 3D MRA dataset. We take IXI dataset as an example to show the effectiveness of the proposed cropping strategy. We compute the cropping rate (CR) for each subject as the following formula:
\begin{equation}
    \mathrm{CR} = \frac{H\cdot W\cdot D - H^{'}\cdot W^{'}\cdot D^{'}}{H\cdot W\cdot D}
\end{equation}
where $(H, W, D)$ is the original image size and $(H^{'}\cdot W^{'}\cdot D^{'})$ is the image size after cropping. Higher CR represents more background pixels being removed. The CR of the IXI dataset is shown in Fig. \ref{fig:cropping effectiveness}. The average CR is 42.99\%, and 80\% of the samples subtracted nearly 40\% (38.49\%) of the background points. Besides, two representative data are shown in Fig. \ref{fig:cropping view}. The strategy of non-zero cropping is widely employed in the analysis of 3D medical images. Yet, the presence of noise in MRI scans undermines the effectiveness of this approach, leading to suboptimal image cropping. Our proposed methodology addresses this issue by meticulously eliminating the maximum amount of background pixels. This approach gains importance in scenarios involving extensive 3D datasets, ensuring a more precise and efficient analysis.

Our proposed method is not limited by the backbone structure and has relatively high generalizability. More importantly, the proposed pretraining strategy may reduce the performance gap among the various backbone models. It is observed from Table \ref{tab: backbone} that the performance gap between Swin UNETR and UNet is around 6.5 before pretraining, while it becomes approximately 2 after using the proposed pretraining strategy. This indicates that the lightweight model (e.g., vanilla UNet) can also exhibit competitive performance after employing the proposed pretraining method, which holds significant implications for the use of lightweight networks in clinical settings.

One limitation of this study is that our method may not effectively use the other modal data that can not fully reflect the vessel morphology (e.g., T1). The proposed pretraining method relies on the preprocessing workflow for cerebrovascular extraction, while other modal data might not include any vessel-related information. Any modality of data, such as computed tomographic angiography, digital subtraction angiography, and optical coherence tomography, capable of capturing vessel morphology, can derive advantages from the proposed method. Besides, our pretraining method can serve as an auxiliary training approach for a SSL model trained using other modal data or general large vision models, and help to construct a domain-specific model for cerebrovascular segmentation.

\section{Conclusion}
In this study, we introduce a Frangi filtering-based pretraining network to effectively leverage the unlabeled TOF-MRA data. Our method capitalizes on the prior knowledge of tubular structures by incorporating it into the pretraining tasks, and shows great superiority over other SemiSL and SSL methods on four cerebrovascular segmentation datasets. The ablation studies also show that our proposed method significantly improves model performance, particularly with a limited number of labeled data. Besides, our method is not restricted by model architecture and can enhance the performance of various backbone structures.

\section*{Acknowledgments}
This work was supported in part by the National Natural Science Foundation of China under Grant: 62027901, 81930053, 81227901; Beijing Natural Science Foundation: JQ22023; CAS Youth Innovation Promotion Association under Grant Y2022055. The authors would like to acknowledge the instrumental and technical support of Multimodal Biomedical Imaging Experimental Platform, Institute of Automation, Chinese Academy of Sciences.

\bibliographystyle{model2-names.bst}\biboptions{authoryear}
\bibliography{refs}

\clearpage
\section*{Supplementary Material}
\subsection{Pretraining Datasets}
The pretraining dataset consists of 5 publicly accessible datasets:
\begin{itemize}
    \item \textbf{IXI}\footnote{\url{http://brain-development.org/ixi-dataset/}}: This dataset contains 570 subjects. 45 subjects with vessel masks \cite{chen2022attention} are used for evaluation dataset. The rest 525 subjects are used for pretraining. The image size of most data (497 subjects) is (512, 512, 100). The image size of 27 subjects is (1024, 1024, 92) and one subject is (1024, 1024, 91). The spacing is (0.4688, 0.4688, 0.8000).
    \item \textbf{OASIS}\footnote{\url{https://www.oasis-brains.org}}: This dataset contains 525 subjects. The spacing and image size are not identical among the subjects. The average image size is (583, 765, 216). The spacing of most data (337 subjects) is (0.2995, 0.2995, 0.6000), and the average spacing is (0.2977, 0.2977, 0.5933).
    \item \textbf{TubeTK}\footnote{\url{https://public.kitware.com/Wiki/TubeTK/Data}}: This dataset contains 109 subjects. 42 subjects with vessel masks are used for the evaluation dataset. The rest 67 subjects are used for pretraining. The image size is (448, 448, 128) and the spacing is (0.5134, 0.5134, 0.8000).
    \item \textbf{ADAM}\footnote{\url{https://adam.isi.uu.nl/data/}}: This dataset contains 113 subjects. Each subject contains one original MRA-TOF image and one bias field corrected MRA-TOF image. Only the original data are used in this study. The average image size is (556, 556, 131). The spacing is various and the average spacing is (0.3524, 0.3524, 0.5447).
    \item \textbf{BrainAneurysm}\footnote{\url{https://openneuro.org/datasets/ds003949/versions/1.0.1}}: This dataset contains 284 subjects, of which 127 are healthy controls and 157 are patients with brain aneurysms. Four subjects (sub-200, sub-235, sub-315 and sub-450) are removed due to the incomplete image data. The average spacing is (0.4021, 0.4021, 0.6613), and the average image size is (466, 546, 147).
\end{itemize}

\subsection{Evaluation Datasets}
The raw image data of the three labeled datasets are accessible at the following public websites:
\begin{itemize}
    \item \textbf{TubeTK-42}: \url{https://public.kitware.com/Wiki/TubeTK/Data}
    \item \textbf{Brains}: \url{https://www.nitrc.org/projects/icbmmra/}
    \item \textbf{EDEN}: \url{https://zenodo.org/record/3994749}
\end{itemize}
We provide the voxel-wise vessel mask for the three datasets in this study.

\subsection{The implementations of baseline models}
For the baseline models, we use the following publicly available implementations.
\begin{itemize}
    \item \textbf{GCS}: \url{https://github.com/MontaEllis/SSL-For-Medical-Segmentation}
    \item \textbf{DTC}: \url{https://github.com/HiLab-git/DTC} 
    \item \textbf{GBDL}: \url{https://github.com/Jianf-Wang/GBDL}
    \item \textbf{SSNet}: \url{https://github.com/ycwu1997/SS-Net} 
    \item \textbf{BCP}: \url{https://github.com/DeepMed-Lab-ECNU/BCP}
    \item \textbf{MAE}: \url{https://github.com/facebookresearch/mae}
    \item \textbf{Swin UNETR}: \url{https://github.com/Project-MONAI/research-contributions/tree/main/Swin UNETRR} 
    \item \textbf{UniMiSS}: \url{https://github.com/YtongXie/UniMiSS-code}
    \item \textbf{GVSL}: \url{https://github.com/YutingHe-list/GVSL} 
    \item \textbf{PCRL}: \url{https://github.com/Luchixiang/PCRL}
\end{itemize}
The UNet is implemented in MONAI\footnote{https://monai.io/}.

\end{document}